\pdfoutput=1
\documentclass[11pt,table,xcdraw]{article}

\usepackage{EACL2023}

\usepackage{times}
\usepackage{latexsym}

\usepackage[LAE, T5,T1]{fontenc} % LAE for Arabic, T5 for Vietnamese, T1 for Icelandic

\usepackage[utf8]{inputenc}

\usepackage{microtype}

\usepackage{inconsolata}

\usepackage{graphicx}
\usepackage{booktabs}
\usepackage{multirow}
\usepackage{adjustbox} % resize or center tables
\usepackage{xcolor}
\usepackage{placeins}

\newlength{\spacer}
\newcommand{\token}[1]{{\small\texttt{#1}}}
\newcommand{\hash}{{\footnotesize\texttt{\#\#}}}
\newcommand{\minispace}{\hspace{5pt}}
\newcommand{\reduceskip}{\vspace{-5pt}}
\newcommand{\largerskip}{\vspace{6pt}}
\newcommand{\noise}[1]{\textcolor{red}{\textbf{#1}}}

\newcommand*{\transfer}{$\rightarrow$}

\newcommand*{\linkformat}[1]{\texttt{\color{darkblue}#1}}

\newcommand*{\https}[2]{\href{https://#1}{\linkformat{#2}}}
\newcommand*{\udurl}[2]{\href{https://github.com/UniversalDependencies/UD_#1-#2}{\linkformat{github.com/\allowbreak{}Universal\allowbreak{}Dependencies/\allowbreak{}UD\_#1-#2}}}

\title{Does Manipulating Tokenization Aid Cross-Lingual Transfer?\\ A Study on POS Tagging for Non-Standardized Languages}

\author{Verena Blaschke\\
\\
\\
{\tt blaschke@cis.lmu.de}
\And Hinrich Schütze\\
  Center for Information and Language Processing (CIS), LMU Munich, Germany\\
  Munich Center for Machine Learning (MCML), Munich, Germany\\
  {\tt inquiries@cislmu.org}
\And Barbara Plank\\
\\
\\
  {\tt bplank@cis.lmu.de}
  }

\begin{document}
\maketitle

\begin{abstract}
One of the challenges with finetuning pretrained language models (PLMs) is that their tokenizer is optimized for the language(s) it was pretrained on, but brittle when it comes to previously unseen variations in the data. 
This can for instance be observed when finetuning PLMs on one language and evaluating them on data in a closely related language variety with no standardized orthography. 
Despite the high linguistic similarity, tokenization no longer corresponds to meaningful representations of the target data, leading to low performance in, e.g., part-of-speech tagging. 

In this work, we finetune PLMs on seven languages from three different families and analyze their zero-shot performance on closely related, non-standardized varieties. 
We consider different measures for the divergence in the tokenization of the source and target data, and the way they can be adjusted by manipulating the tokenization during the finetuning step. 
Overall, we find that the similarity between the percentage of words that get split into subwords in the source and target data (the \textit{split word ratio difference}) is the strongest predictor for model performance on target data.

\end{abstract}

\section{Introduction}

Transformer-based pre-trained language models (PLMs) enable successful cross-lingual transfer for many natural language processing tasks. However, the impact of tokenization and its interplay with transferability across languages, especially under-resourced variants with no orthography, has obtained limited focus so far.  Tokenization splits words into subwords, but not necessarily in a meaningful way.
An example with a current PLM is illustrated for Alsatian German in Figure~\ref{fig:example}a.
This problem is especially pronounced for vernacular languages and dialects, %variants without standard orthography, 
where words tend to be split at a much higher rate than the standard. This has been observed on, e.g.,
informally written Algerian Arabic \cite{touileb-barnes-2021-interplay}.
As poor subword tokenization can lead to suboptimal language representations and impoverished transfer, it becomes important to understand if the effect holds at a larger scale.
We are particularly interested in challenging setups in which, despite \textit{high language similarity,} comparatively low transfer performance is obtained.

\begin{figure}[t]
\begin{adjustbox}{max width=\textwidth, center}
\begin{tabular}{@{}l@{\minispace}l@{\minispace}l@{\minispace}l@{\minispace}l@{\minispace}l@{}}
\textit{a.}&M'r & redd & alemànnischi & Mundàrte & . \\[-3pt]
&\token{M}, \token{'}, \token{r} & \token{red}, \hash\token{d} & \token{al}, \hash\token{em}, \token{\#\#à,} & \token{Mund,} & \token{.}\reduceskip\\
&&&\token{\#\#nn}, \token{\#\#isch}, \token{\#\#i}& \token{\#\#à}, \token{\#\#rte}&\largerskip\\
\textit{b.}&Wir & sprechen & alemannische & Mundarten & . \\[-3pt]
&\token{Wir} & \token{sprechen} & \token{al}, \token{\#emann}, & \token{Mund}, \token{\#\#arten} & \token{.}\reduceskip\\
&&& \token{\#\#ische}&&\largerskip\\
\textit{c.}&W\noise{(}r & sprechen & alema\noise{I}nische & Mundarten & . \\[-3pt]
&\token{W}, \token{(}, \token{r} & \token{sprechen} & \token{al}, \hash\token{ema}, \token{\#\#In,} & \token{Mund}, \token{\#\#arten} & \token{.} \reduceskip\\
&&&\token{\#\#ische}&&\\
\end{tabular}
\end{adjustbox}
\caption{``We speak Alemannic dialects'', tokenized by GBERT.
Compared to Standard German~(\textit{b.}), the quality of the Alsatian German~(\textit{a.}) tokenization is poor, making cross-lingual transfer hard.
Noise injection~(\textit{c.}) often improves transfer from standard to poorly tokenized non-standardized varieties.}

\label{fig:example}
\end{figure}

A recent study proposes an elegant and lean solution to address this `tokenization gap,' without requiring expensive PLM re-training: to \textit{manipulate tokenization} of PLMs post-hoc~\cite{aepli-sennrich-2022-improving}, i.e., during finetuning by injecting character-level noise (Figure~\ref{fig:example}c).
Noise injection has been shown to successfully aid cross-lingual transfer and is an appealing solution, as it is cheap and widely applicable. 
In this work, we first provide a reproduction study and then broaden it by a systematic investigation of the extent to which noise injection helps.
We also show how it influences the subword tokenization of the source data vis-à-vis the target data. 
We hypothesize that, while not emulating dialect text, injecting noise into standard language data can raise the tokenization rate to a similar level, which aids transfer.

The importance of token overlap between source and target is an on-going debate (to which we contribute):
Prior research has found that subword token overlap between the finetuning and target language improves transfer \cite{wu-dredze-2019-beto, pires-etal-2019-multilingual}, although it might neither be the most important factor \cite{k2020mbert-analysis, muller2022languages} nor a necessary condition for cross-lingual transfer to work \cite{pires-etal-2019-multilingual, conneau-etal-2020-emerging}.

To enable research in this direction, we contribute a novel benchmark. 
We collected under-resourced language variants covering seven part-of-speech (POS) tagging transfer scenarios within three language families. 
This collection enables also future work to study cross-lingual and cross-dialect transfer.

Our contributions are:
\begin{itemize}
    \item We investigate the noise injection method by \citet{aepli-sennrich-2022-improving} with respect to the ideal noise injection rate for different languages and PLMs.
    \item To the best of our knowledge, this is the broadest study that focuses specifically on transfer to closely related, non-standardized language varieties with languages from multiple linguistic families.
    We convert several dialect datasets into a shared tagset (UPOS) and share the conversion scripts.
    \item We compare the effect of noise injection on the subword tokenization differences between the source and target data, and the effect of these differences on the model performance, and find that
    the proportions of (un)split words are a better predictor than the ratio of seen subword tokens.
\end{itemize}

\section{Method}

We make our code, including scripts for reproducing the benchmark, available at \href{https://github.com/mainlp/noisydialect}{\tt github.com/\allowbreak{}mainlp/\allowbreak{}noisydialect}.

\subsection{Injecting Character-Level Noise}

We follow the approach by \citet{aepli-sennrich-2022-improving} to add noise to the finetuning datasets.
Given a noise level $0 \leq n \leq 1$ and a finetuning dataset $F$ with a grapheme inventory $\mathcal{I}$,\footnote{%
    Unlike \citet{aepli-sennrich-2022-improving}, we also include non-alphabetic characters in $\mathcal{I}$, as some of the orthographic differences are punctuation-based (see Figure~\ref{fig:example}).}
we inject noise into each sentence $S \in F$ as follows:
we randomly select $n|S|$ words,\footnote{Excluding words that only contain numerals or punctuation marks.} and for each of these words, we randomly perform one of the three following actions:
\begin{itemize}
    \item delete one randomly chosen character
    \item replace one randomly chosen character with a random character $\in \mathcal{I}$
    \item insert one random character $\in \mathcal{I}$ into a random slot within the word.
\end{itemize}

\citet{aepli-sennrich-2022-improving} investigate transferring POS tagging models to five target languages (Swiss German, Faroese, Old French, Livvi and Karelian) and compare set-ups with no noise (${n=0}$) to adding noise with ${n=0.15}$.
They find that, when the source and target languages are closely related, the configuration with noise consistently performs better. 
We additionally experiment with adding noise at higher levels: to 35~\%, 55~\%, 75~\% and 95~\% of each sentence's tokens.

%\subsection{Subword tokenization measures to quantify data shift}
\subsection{Comparing Datasets via Subword Tokenization}
\label{sec:measures}

We consider several simple measures of comparing the subword tokenization of the source data with that of the target data:

\begin{itemize}
    \item \textit{Split word ratio difference:} The (absolute) difference between the ratios of words that were split into subword tokens in the source and target data.
    (We additionally considered the average number of subword tokens per word, but found that that measure yielded very similar results to the split word ratio difference.)
    \item \textit{Seen subwords} and \textit{seen words:} The ratios of the target subword tokens and target words,\footnote{We consider words here as the annotated units provided by the datasets.} respectively, that are also in the source data.
    (We also included type-based versions of these measures, but found that they behaved similarly to their token-based counterparts.)
    \item \textit{Type--token ratio (TTR) ratio:} The subword-level type-token ratio of the target data divided by that of the source data. This is similar to the TTR-based measures used by \citet{lin-etal-2019-choosing} and \citet{muller2022languages}. 
\end{itemize}

\section{Experimental Set-up}
\subsection{Data}

We analyze transfer between eight source and 18 target datasets in the following language varieties (see Appendix~\ref{sec:appendix-data} for details):
\begin{itemize}
    \item Modern Standard Arabic (MSA) \cite{padt} \transfer{} Egyptian, Levantine, Gulf and Maghrebi Arabic \cite{darwish-etal-2018-multi}
    \item German \cite{borges-volker-etal-2019-hdt} \transfer{} Swiss German \cite{hollenstein2014compilation}, Alsatian German \cite{bernhard2019alsatian}
    \item German \cite{borges-volker-etal-2019-hdt}, Dutch \cite{bouma-van-noord-2017-increasing} \transfer{} Low Saxon \cite{siewert-etal-2021-towards-fixed}
    \item Norwegian (Nynorsk) \cite{velldal-etal-2017-joint}, Norwegian (Bokmål) \cite{ovrelid-hohle-2016-universal} \transfer{} West, East and North Norwegian \cite{ovrelid-etal-2018-lia}
    \item French \cite{guillaume2019conversion} \transfer{} Picard \cite{martin2018picard}
    \item French \cite{guillaume2019conversion}, Spanish \cite{taule-etal-2008-ancora} \transfer{} Occitan \cite{bras2018occitan}
    \item Finnish \cite{pyysalo2015udfinnish} \transfer{} six Finnish dialect groups \cite{LA-murre-vrt_en}
\end{itemize}

This list includes varieties from three language families (Afro-Asiatic, Finno-Ugric and Indo-European), written in two types of writing systems (alphabetical and abjad). 
% The non-standardized datasets also differ in the kinds of data they contain---for instance tweets, Wikipedia articles or professionally transcribed interviews.
It also covers a range of different degrees of linguistic relatedness (e.g., the Norwegian dialects are much more closely related to each other and to the standardized varieties than can be said of the Arabic group) and text genres (including tweets, Wikipedia articles, and professionally transcribed interviews).
While orthographies for some of our target languages (e.g., Low Saxon) have been proposed, none of these languages have a sole orthography that is used by virtually all speakers.

Many of these corpora are from the Universal Dependencies (UD) project \cite{zeman2022ud-2-11}, or annotated according to UD's POS tagging scheme (UPOS).
For some language varieties, we first make the data compatible with UPOS:
We convert the tagsets used for the Arabic dialects and the Finnish dialects to UPOS (Appendix~\ref{sec:conversion}).
To process the Occitan data, we separate contractions (\textsc{adp+det}), similarly to the way these cases are handled in other Romance UD treebanks.\footnote{\raggedright E.g., \https{universaldependencies.org/fr/tokenization.html}{universaldependencies.org/fr/\allowbreak{}tokenization.\allowbreak{}html}}
For the Norwegian dialects, we merge parallel data from the original corpus (dialect vs.\ orthographic transcriptions) with the orthography-only treebank to get a treebank with dialect transcriptions.%
\footnote{\raggedright The resulting scripts are available at \texttt{github.com/mainlp/}\allowbreak{}\{\allowbreak{}\href{https://github.com/mainlp/convert-qcri-4dialects}{\texttt{convert-qcri-4dialects}}, \href{https://github.com/mainlp/convert-la-murre}{\texttt{convert-la-murre}}, \href{https://github.com/mainlp/convert-restaure-occitan}{\texttt{convert-restaure-occitan}}, \href{https://github.com/mainlp/UD_Norwegian-NynorskLIA_dialect}{\texttt{UD\_Norwegian-NynorskLIA\_dialect}}\}.}

\subsection{PLMs}

We use two multilingual PLMs: mBERT \cite{devlin-etal-2019-bert} and XLM-R \cite{conneau-etal-2020-unsupervised}.
Additionally, we include one monolingual model per source language.
Both multilingual PLMs included all of our source languages in their pretraining data, and mBERT also contains two of our target languages (Low Saxon and Occitan).
Details on the PLMs we used can be found in Appendix~\ref{sec:appendix-plms}.

We use \textit{base}-size, cased versions of all models, and finetune the PLMs on the default training data subsets.
We perform a simple grid search to choose one set of hyperparameters to be used for all experiments.
This grid search was performed on the German (and Swiss German), Arabic (and Egyptian), and Finnish (and Savonian Finnish) data, using XLM-R and the respective monolingual models. 
Table~\ref{tab:hyperparam} in Appendix~\ref{sec:appendix-plms} contains details on the hyperparameters.

\section{Results and Discussion}

\begin{table*}[p]
\begin{adjustbox}{max width=\textwidth, center}
% workaround for arxiv
% \input{tables/accuracies.tex}
\includegraphics[width=\textwidth]{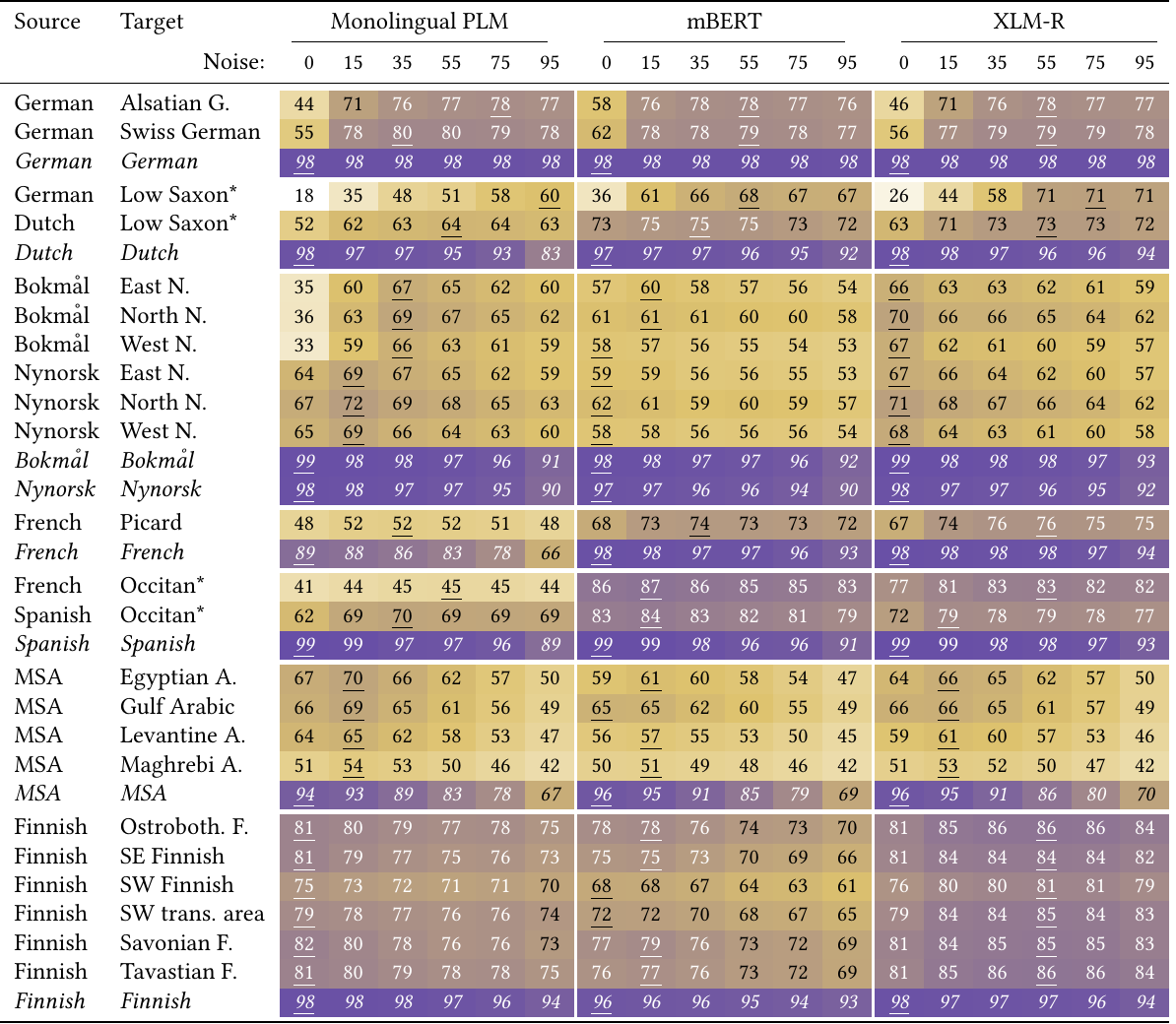}
\end{adjustbox}
\caption{\textbf{Accuracy scores (in~\%) by language combination, language model and noise level.} 
Scores are averaged over five initializations.
Target languages marked with an asterisk* appear in the training data for mBERT.
Rows \textit{in italics} contain scores on the test splits of the datasets used for finetuning.
The best accuracy for each language pair and PLM combination is \underline{underlined}.
}
\label{tab:accuracy}
\end{table*}

All results we report are averaged over five different random initializations.
Table~\ref{tab:accuracy} shows the accuracy scores of the inferred POS tags.
We observe similar trends for the macro-averaged F1 score as well.

\paragraph{Zero-shot transfer.}
Performance on the unseen test languages/dialects is much lower than on the test partitions of the corpora on which the models were finetuned.
This is expected, as there are not only orthographic and stylistic differences between the corpora, but also some grammatical differences between the language varieties.

The extent to which performance drops is language-dependent:
For instance, the best results for the Finnish dialects are 12--17 percentage points below the best results for the Finnish standard language (XLM-R), whereas the best results for the Norwegian dialects are 26--29 percentage points below the standard language accuracy (Nynorsk with NorBERT).
When we have multiple target dialects for one source language, the target scores tend to be similar to one another across noise levels and PLM choices.

\paragraph{PLM choice matters for low-resource languages.}
While the models are for the most part indistinguishable in their performance on the source languages, the performance on the target languages can vary substantially.
For instance, XLM-R outperforms mBERT and FinBERT on the Finnish dialect data.
Similarly, both multilingual models perform much better than the monolingual models on the Low Saxon, Picard and Occitan data, and the reverse is true for the Arabic dialects.
Neither the performance on the source languages nor the transfer performance with $n=0$ reveal which model performs best on the target data when the ideal amount of noise is added.

\paragraph{Effect of noise level on accuracy.}
The optimal noise level depends on the language pair and on the PLM -- there is no universal best noise level choice.
In many (but not all) cases, the accuracy rises drastically when increasing the noise level from 0\,\% to 15\,\%, and the (positive or negative) differences between subsequent noise levels are less pronounced.
The noise level of 15\,\% used by \citet{aepli-sennrich-2022-improving} is thus a reasonable choice, although not always optimal.
In some cases, the accuracy might be much greater at a different noise level (e.g., in the German\transfer{}Low German XLM-R set-up the maximum gain compared to using no noise is +42~percentage points; +27~compared to 15\,\% noise).
In other cases, adding any noise at all decreases the performance -- most drastically in the case of Bokmål\transfer{}West Norwegian with XLM-R, where the accuracy drops by 5~percentage points when using 15\,\% noise instead of no noise at all.
However, the general trend is that accuracy as a function of noise has a single global maximum and no local maxima -- there is a clear optimum level of noise in almost all cases.\footnote{%
The minor exceptions to this are FinBERT's performance on the Ostrobothnian and South-East Finnish data and XLM-R's predictions for the Spanish\transfer{}Occitan transfer (see Table~\ref{tab:accuracy}).
In all of these cases, a second increase occurs after the maximum accuracy has already been reached and stays below this maximum.}

Performance on the standard language test splits from the corpora used for finetuning always decreases when noise is introduced.
Whether this is detrimental depends on the language: the accuracy on the German test set only drops very slightly (less than one percentage point) whereas the quality of the tag predictions for MSA deteriorates considerably, independently of the model used. % (up to -26 or -27 percentage points for each model). 

\begin{figure}[t]
    \centering
    % trim: left, bottom, right, top
    \includegraphics[width=\columnwidth, clip, trim={2.8mm 13mm 6mm 14mm}]{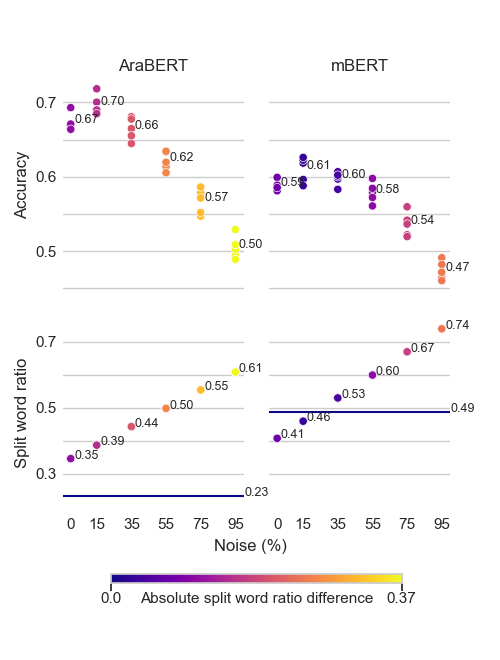}
    \caption{\textbf{Transfer from MSA to Egyptian Arabic with AraBERT (left) and mBERT (right).}\\
    \textbf{Top:} Accuracy scores per language model and noise level (five initializations per set-up; the numbers in the scatterplot indicate the mean accuracy per set-up).\\
    \textbf{Bottom:} Split word ratios per language model and noise level for the source data (dots) and the target data (dark blue lines) (five initializations per set-up).
    The colours indicate the (absolute) difference between the split word ratio of the training and target data (darker~=~smaller difference).
    }
    \label{fig:splittok-egyptian}
\end{figure}

\paragraph{Effect of noise level on split word ratio difference.}
The words in the target data tend to be split into subword tokens more often than is the case for the source data.\footnote{%
The exceptions to this are the tokenization of the Finnish dialects by the multilingual models and the tokenization of the Arabic dialects with AraBERT.
The latter is likely due to AraBERT including a pre-tokenization step that splits words into stems and affixes \cite{antoun-etal-2020-arabert}, 
but MSA and non-standard varieties of Arabic having morphological differences.
}
Increasing the noise level during finetuning results in the source data being split into more subword tokens (see the rising sequences of dots in the lower part of Figure~\ref{fig:splittok-egyptian}).
In all set-ups, the split word ratio of the source data is higher than that of the target data when $n\geq0.75$.

\begin{table}[t]
\begin{adjustbox}{max width=\columnwidth, center}
% workaround for arxiv
% \input{tables/correlations.tex}
\includegraphics[width=\textwidth]{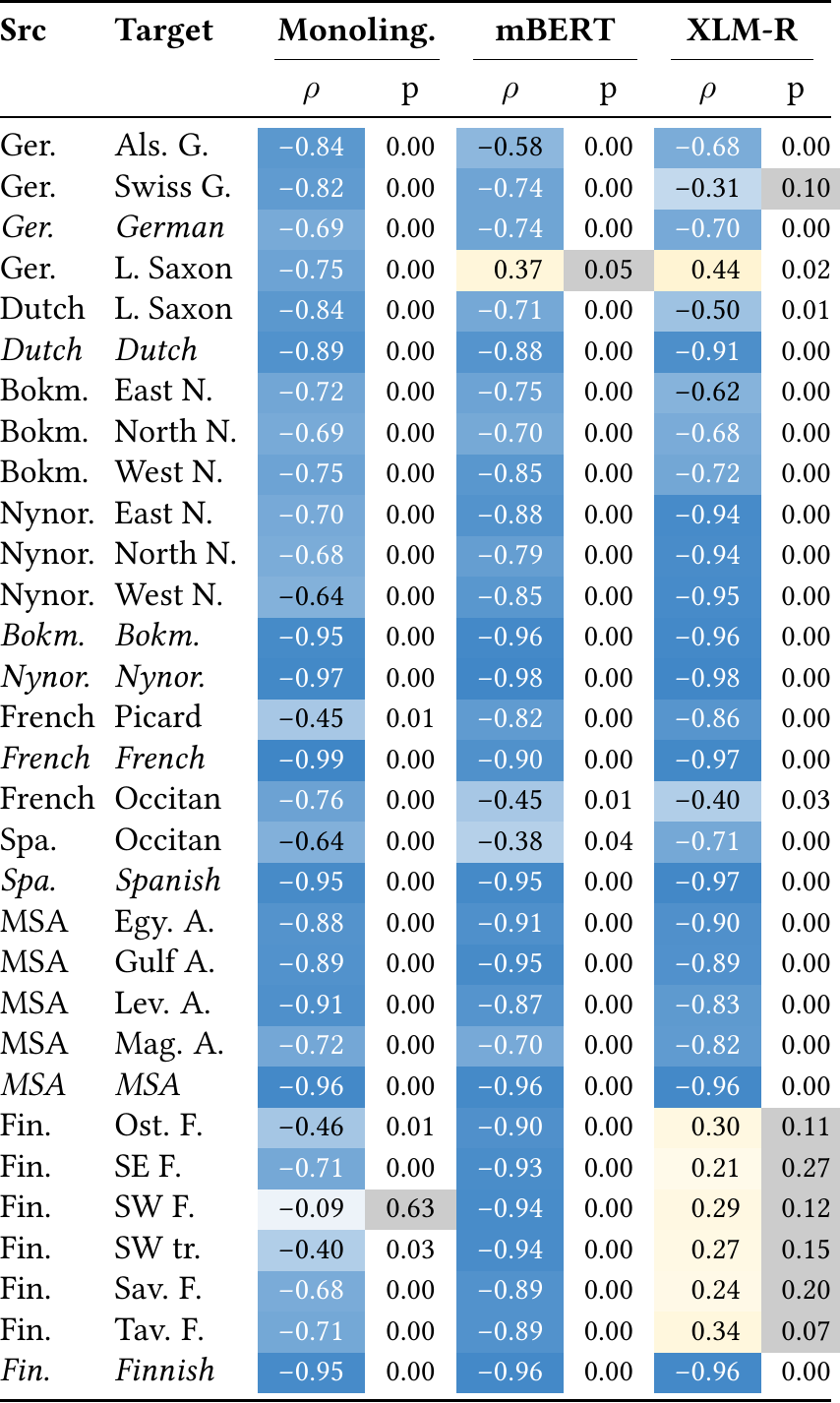}
\end{adjustbox}
\caption{\textbf{Correlation between split word ratio difference and accuracy.}
Spearman's $\rho$ with \textit{p}-values for all noise levels and random initializations per language pair and PLM.
Negative correlations are highlighted in blue, positive ones in yellow.
\textit{P}-values of 0.05 and above have a grey background.
}
\label{tab:correlation}
\end{table}

\paragraph{Effect of split word ratio difference on accuracy.}
Out of the subword tokenization measures introduced in Section~\ref{sec:measures}, the \textit{split word ratio difference} correlates most consistently with the performance: the smaller the difference is (i.e., the more similar the ratios are), the higher the accuracy tends to be (Table~\ref{tab:correlation}).
Figure~\ref{fig:splittok-egyptian} shows an example; note that the correlation is stronger for the model on the right-hand side (mBERT) than for the model on the left (AraBERT).

The correlation is strong enough that, {if one really wants
to avoid including} the noise level in a hyperparameter
search, only carrying out the cheap calculations needed for
the \textit{split word ratio difference} and choosing the
noise level with the lowest difference can be a proxy.
Nevertheless, the correlation is not perfect and this method does not necessarily  pick the best noise level.

\subsection{Additional Findings}

\paragraph{The role of seen (sub)words.}

Adding noise to the source data initially increases the word and subword token overlap with the target data for all cross-lingual/cross-dialectal set-ups, regardless of model choice.
As the noise level increases, this trend ultimately reverses, although the source and target data still have a greater (sub)word overlap at $n=0.95$ than at $n=0$.

The seen word ratio and seen subword ratio are much poorer predictors for the model performance than the split word ratio difference is.
They are much less consistent and correlate positively with accuracy for many set-ups but negatively for many others, and the correlations tend to have larger \textit{p}-values (see Tables~\ref{tab:corr-seen-subwords} and~\ref{tab:corr-seen-words} in Appendix~\ref{sec:appendix-correlations} for details).
While prior works have come to conflicting conclusions regarding the importance of subword token overlap for transfer between more distantly related (or unrelated) languages \cite{wu-dredze-2019-beto, pires-etal-2019-multilingual, k2020mbert-analysis, conneau-etal-2020-emerging, muller2022languages}, 
we find that it is a very poor predictor for the transfer between very closely related languages when injecting character-level noise.
One possibility for this is that the seen target subwords contained in the noisy source data might not necessarily belong to the same POS classes.

\paragraph{The role of TTR ratio.}
For most set-ups, the TTR ratio initially decreases before ultimately increasing, with no local minima.
In all of our experiments, the TTR ratio either always stays above one (the target data's TTR remains higher than that of the source data) or always below one (the source data's TTR stays higher than that of the target data; this is only the case for the cross-dialected Finnish set-ups) -- adding noise does not result in bringing the TTRs to a similar level.
The TTR ratio correlates positively with accuracy for some set-ups and negatively with others (see Table~\ref{tab:corr-ttr-ratio} in Appendix~\ref{sec:appendix-correlations}).
This overall very weak {predictive capacity} of the TTR ratio is similar to what \citet{muller2022languages} find for named entity recognition and in line with \citeauthor{lin-etal-2019-choosing}'s (\citeyear{lin-etal-2019-choosing}) results for POS tagging -- their TTR-based measure is only a useful performance predictor when used in conjunction with other measures.

\section{Conclusion}

We have confirmed the usefulness of the noise injection method by \citet{aepli-sennrich-2022-improving} for model transfer between closely related languages.
To that end, we have converted additional dialectal datasets to the UPOS standard and make the conversion code available to other researchers.
Furthermore, we have shown that the ideal amount of noise that should be injected at finetuning time depends on the languages and PLMs used.
We have also investigated the role that subword tokenization
plays in this and found that the \textit{split word ratio
difference} -- the (absolute) difference between the
proportion of words split into subword tokens in the source
and target data -- is
a reliable, albeit imperfect,
predictor of the performance of the transfer model.

\section*{Limitations}

We include data from three linguistic families, as we were not able to find additional accessible high-quality dialect datasets manually annotated with POS tags for more linguistic families.
This general lack of annotated resources is also why we were only able to focus on one NLP task.
The tagsets for the Arabic and Finnish varieties were converted to UPOS by a linguist who is not a specialist of Arabic or Finnish.

We only consider one way of modifying the tokenization.
In future research, it would be interesting to also consider BPE dropout \cite{provilkov-etal-2020-bpe}, which \citet{aepli-sennrich-2022-improving} show to have an effect on transfer between related languages that is somewhat similar to that of noise injection.
It would also be of interest to investigate token-free models like ByT5 \cite{xue-etal-2022-byt5} or CharacterBERT \cite{el-boukkouri-etal-2020-characterbert}, the latter of which has proven useful for processing data in a non-standard variety of Arabic \cite{riabi-etal-2021-character}.

\section*{Acknowledgements}

We thank the members of the MaiNLP research group as well as the anonymous reviewers for their useful feedback. 
This research is supported by European Research Council (ERC) Consolidator Grant DIALECT 101043235.
This work was partially funded by the ERC under the European Union's Horizon 2020 research and innovation program (grant 740516).

\FloatBarrier

\bibliography{anthology,custom,ud}
\bibliographystyle{acl_natbib}

\appendix

\section{Dataset Details}
\label{sec:appendix-data}

These are the datasets we use in this study:

\begin{itemize}
\item  Modern Standard Arabic: UD Arabic PADT \cite{padt} --  CC BY-NC-SA~3.0 -- \udurl{Arabic}{PADT}
\item  Egyptian, Levantine, Gulf and Maghrebi Arabic: QCRI Dialectal Arabic Resources \cite{darwish-etal-2018-multi}  --  Apache License~2.0 -- \https{alt.qcri.org/resources/da\_resources/}{alt.qcri.org/resources/da\_resources}
\item German: UD German HDT \cite{borges-volker-etal-2019-hdt,foth2014hdt}  --  CC BY-SA~4.0 -- \udurl{German}{HDT}
\item Swiss German: NOAH v~3.0 (UPOS-tagged subset) \cite{hollenstein2014compilation,aepli-sennrich-2022-improving}  --  CC BY~4.0 -- \https{github.com/noe-eva/NOAH-Corpus}{github.com/noe-eva/NOAH-Corpus}
\item Alsatian German: Annotated Corpus for the Alsatian Dialects \cite{bernhard2019alsatian,bernhard-etal-2018-corpora}  --  CC BY-SA~4.0 -- \https{zenodo.org/record/2536041}{zenodo.org/\allowbreak{}record/\allowbreak{}2536041}. 
Like Swiss German, Alsatian German is a variety of Alemannic German.
Note that while both NOAH and the Alsatian corpus contain parts of the Alemannic Wikipedia, the corpora do not overlap.
\item Dutch: UD Dutch Alpino \cite{bouma-van-noord-2017-increasing, vanderbeek2002alpino} --  CC BY-SA~4.0 -- \udurl{Dutch}{Alpino}
\item Low Saxon: UD Low Saxon LSDC \cite{siewert-etal-2021-towards-fixed}  --  CC BY-SA~4.0 -- \udurl{Low\_Saxon}{LSDC}
\item Norwegian (Nynorsk): UD Norwegian Nynorsk \cite{velldal-etal-2017-joint,solberg-etal-2014-norwegian}  --  CC BY-SA~4.0 -- \udurl{Norwegian}{Nynorsk}
\item Norwegian (Bokmål): UD Norwegian Bokmaal \cite{ovrelid-hohle-2016-universal,solberg-etal-2014-norwegian}  --  CC BY-SA~4.0 -- \udurl{Norwegian}{Bokmaal}
\item West, East and North Norwegian: dialect transcriptions: LIA Norwegian---Corpus of historical dialect recordings \cite{ovrelid-etal-2018-lia} --  CC BY-NC-SA~4.0 -- \https{tekstlab.uio.no/LIA/norsk/}{tekstlab.uio.no/LIA/norsk}; treebank: UD Norwegian NynorskLIA  \cite{ovrelid-etal-2018-lia} --  CC BY-SA~4.0 -- \udurl{Norwegian}{NynorskLIA}. 
The Trønder data (Lierne/Nordli) from the same dataset are omitted because their sample size is much smaller than those of the other dialect groups.
We group the remaining locations as follows: East Norwegian (Ål, Bardu,\footnote{The history of the dialects spoken in and around Bardu is complex, as it is a contact point of East and North Norwegian. For more information, see \citet{jahr1996indre-troms}.} Eidsberg, Gol), West Norwegian (Austevoll, Farsund/Lista, Giske), North Norwegian (Flakstad, Vardø).
\item French: UD French GSD \cite{guillaume2019conversion}  --  CC BY-SA~4.0 -- \udurl{French}{GSD}
\item Picard: Annotated Corpus for Picard \cite{martin2018picard,bernhard-etal-2018-corpora}  -- CC BY-SA~4.0 -- \https{zenodo.org/record/1485988}{zenodo.org/record/1485988}
\item Spanish: UD Spanish AnCora \cite{taule-etal-2008-ancora}  --  CC BY~4.0 -- \udurl{Spanish}{AnCora}
\item Occitan: Annotated Corpus for Occitan \cite{bras2018occitan,bernhard-etal-2018-corpora}  -- CC  BY-SA~4.0 -- \https{zenodo.org/record/1182949}{zenodo.org/record/1182949}
\item Finnish: UD Finnish TDT \cite{pyysalo2015udfinnish,haverinen2013tdt}  --  CC BY-SA~4.0 -- \udurl{Finnish}{TDT}
\item Finnish dialects: The Finnish Dialect Corpus of the Syntax Archive, Downloadable VRT Version \cite{LA-murre-vrt_en}  --  CC-BY-ND~4.0 -- \https{urn.fi/urn:nbn:fi:lb-2019092001}{urn.fi/\allowbreak{}urn:nbn:fi:lb-2019092001}. 
We use the dialect regions that are indicated in the corpus: South-Western, South-Eastern, Tavastian, Ostrobothnian, and Savonian dialects, as well as dialects from the transition region between the South-Western area and Tavastia.
\end{itemize}

\section{Tagset Conversion}
\label{sec:conversion}

\subsection{QCRI Dialectal Arabic Resources}
\label{sec:conversion-arabic}

To convert the POS tags of the dialectal Arabic dataset, we use the corpus documentation \cite{darwish-etal-2018-multi}, the documentation of the Farasa tagset \cite{darwish-etal-2014-using} (on which the corpus's tagset is based), the documentation for Arabic treebanks in general and UD Arabic PADT in particular,\footnote{\href{https://universaldependencies.org/ar/index.html}{\tt universaldependencies.org/ar/index.html}; \href{https://universaldependencies.org/treebanks/ar_padt}{\tt universaldependencies.org/treebanks/ar\_padt}}, grammars of standard and non-standard Arabic \cite{ryding2005msa,brustad2000syntax}, and \citeauthor{sanguinetti2022treebanking}'s (\citeyear{sanguinetti2022treebanking}) tagging recommendations for user-generated content.
Table~\ref{tab:arabic-pos} shows how we converted the tags to UPOS.
The \textsc{part} tag is converted to UPOS \textsc{part} unless the associated word form is one of the subordinating conjunctions tagged as such (\textsc{sconj}) in UD Arabic PADT.
Tokens tagged with \textsc{case/nsuff} or \textsc{prog\_part} are fused with preceding \textsc{adj/noun} or \textsc{verb} tokens, when possible. When they appear on their own, they are tagged with \textsc{x}.
Additional tags from the extended Farasa tagset that are not used in the treebank are: \textsc{abbrev}, \textsc{jus}, \textsc{vsuff}.

\begin{table}[t]
\centering
\begin{tabular}{ll}\toprule
UPOS & Farasa (extended) \\\midrule
\textsc{adj} & \textsc{(det+)adj(+case/nsuff)} \\
\textsc{adp} & \textsc{prep} \\
\textsc{adv} & \textsc{adv} \\
\textsc{aux} & \textsc{fut\_part} \\
\textsc{cconj} & \textsc{conj} \\
\textsc{det} & \textsc{det} \\
\textsc{noun} & \textsc{(det+)noun(+case/nsuff)} \\
\textsc{num} & \textsc{num} \\
\textsc{part} & \textsc{part,* neg\_part} \\
\textsc{propn} & \textsc{mention} \\
\textsc{pron} & \textsc{pron} \\
\textsc{punct} & \textsc{punc} \\
\textsc{sconj} & \textsc{part*} \\
\textsc{sym} & \textsc{emot, url} \\
\textsc{verb} & \textsc{(prog\_part+)v} \\
\textsc{x} & \textsc{foreign, hash, case*}\\ 
& \textsc{nsuff,* prog\_part*}\\\bottomrule
\end{tabular}
\caption{\textbf{POS tag conversion for the non-standard Arabic varieties.}
The treatment of tags marked with an asterisk* is explained in the text.
}
\label{tab:arabic-pos}
\end{table}

\subsection{Finnish Dialect Corpus of the Syntax Archive}
\label{sec:conversion-finnish}

The conversion of the Finnish tags is based on documentation for the Finnish Dialect Corpus,\footnote{\href{https://kielipankki.fi/aineistot/la-murre/la-murre-annotaatiot}{\tt kielipankki.fi/aineistot/la-murre/\allowbreak{}la-\allowbreak{}murre-\allowbreak{}annotaatiot}; \href{https://blogs.helsinki.fi/fennistic-info/files/2020/12/2.-Sananmuodot-morfologia-morfosyntaksi.pdf}{\tt blogs.\allowbreak{}helsinki.\allowbreak{}fi/\allowbreak{}fennistic-\allowbreak{}info/\allowbreak{}files/\allowbreak{}2020/\allowbreak{}12/\allowbreak{}2.-\allowbreak{}Sananmuodot-\allowbreak{}morfologia-\allowbreak{}morfo\allowbreak{}syntaksi.\allowbreak{}pdf}} on the UPOS documentation,\footnote{\href{universaldependencies.org/u/pos/all.html}{\tt universaldependencies.org/u/pos/all.html}} and on the documentation of the UD Finnish TDT corpus.\footnote{\href{https://universaldependencies.org/treebanks/fi_tdt}{\tt universaldependencies.org/treebanks/fi\_tdt}}
Table~\ref{tab:finnish-pos} shows the correspondences between the two tagsets.
UD Finnish TDT does not use \textsc{det} or \textsc{part}.
Two tags needed to be further disambiguated: \textit{v} (used for auxiliaries and full verbs) and \textit{q} (used for interrogative words). 
For these entries, we use the lemma to decide which POS a given word belongs to. 

\begin{table}[h]
\centering
\begin{tabular}{ll}\toprule
UPOS & Finnish Dialect Corpus \\\midrule
\textsc{adj} & a, a:pron, a:pron:dem, a:pron:int, \\ 
&a:pron:rel, num:ord, num:ord\_pron, q* \\
\textsc{adp} & p:post, p:pre \\
\textsc{adv} & adv, adv:pron, adv:pron:dem,\\
& adv:pron:int, adv:pron:rel, adv:q, p:adv \\
\textsc{aux} & v*, neg \\
\textsc{cconj} & cnj:coord \\
\textsc{det} & -- \\
\textsc{intj} & intj \\
\textsc{noun} & n \\
\textsc{num} & num:card, num:murto \\
\textsc{part} & -- \\
\textsc{propn} & n:prop, n:prop:pname \\
\textsc{pron} & pron, pron:dem, pron:int, pron:pers,\\
& pron:pers12, pron:ref, pron:rel, q* \\
\textsc{punct} & punct \\
\textsc{sconj} & cnj:rel, cnj:sub \\
\textsc{sym} & -- \\
\textsc{verb} & v* \\
\textsc{x} & muu \\\bottomrule
\end{tabular}
\caption{\textbf{POS tag conversion for the Finnish Dialect Corpus.}
Tags marked with an asterisk* are disambiguated with the help of lexical information.}
\label{tab:finnish-pos}
\end{table}

\section{Language Models}
\label{sec:appendix-plms}

% \todo{present as list or use table?}

We use the following PLMs:

\begin{itemize}
\item mBERT \cite{devlin-etal-2019-bert}\footnote{The article details the architecture. Information on the multilingual version can be found at \https{github.com/google-research/bert/blob/master/multilingual.md}{github.com/\allowbreak{}google-\allowbreak{}research/bert/blob/master/multilingual.md}} -- Apache~2.0 -- \https{huggingface.co/bert-base-multilingual-cased}{huggingface.co/\allowbreak{}bert-\allowbreak{}base-\allowbreak{}multilingual-\allowbreak{}cased}. 
mBERT's pretraining data include all of the source languages from our study.
It also includes Low Saxon and Occitan.
\item XLM-R \cite{conneau-etal-2020-unsupervised} -- MIT licence -- \https{huggingface.co/xlm-roberta-base}{huggingface.co/xlm-roberta-base}.
XLM-R's pretraining data also include all of the source languages from our study.
The documentation does not specify whether the Norwegian pretraining data are written in Bokmål, Nynorsk, or both.
XLM-R was not trained on any of our target languages.
\item Arabic: AraBERT v.~2 \cite{antoun-etal-2020-arabert} -- custom licence\footnote{\https{github.com/aub-mind/arabert/blob/master/arabert/LICENSE}{github.com/aub-mind/arabert/blob/master/\allowbreak{}arabert/LICENSE}} -- \https{huggingface.co/aubmindlab/bert-base-arabertv2}{huggingface.co/\allowbreak{}aubmindlab/bert-base-arabertv2}
\item German: GBERT \cite{chan-etal-2020-germans} -- MIT licence -- \https{huggingface.co/deepset/gbert-base}{huggingface.co/\allowbreak{}deepset/\allowbreak{}gbert-base}
\item Dutch: BERTje \cite{de-vries2019bertje} -- Apache~2.0 -- \https{github.com/wietsedv/bertje}{github.com/\allowbreak{}wietsedv/\allowbreak{}bertje}
\item Norwegian (both Bokmål and Nynorsk): NorBERT v.~2 \cite{kutuzov-etal-2021-large} -- CC0~1.0 -- \https{huggingface.co/ltgoslo/norbert2}{huggingface.co/ltgoslo/norbert2}. 
\item French: CamemBERT \cite{martin-etal-2020-camembert} -- MIT licence -- \https{camembert-model.fr}{camembert-model.fr}
\item Spanish: BETO \cite{canete2020beto} -- CC BY~4.0 -- \https{huggingface.co/dccuchile/bert-base-spanish-wwm-cased}{huggingface.co/\allowbreak{}dccuchile/\allowbreak{}bert-base-spanish-wwm-cased}
\item Finnish: FinBERT v.~1.0 \cite{virtanen2019finbert} -- CC BY~4.0 -- \https{github.com/TurkuNLP/FinBERT}{github.com/\allowbreak{}TurkuNLP/FinBERT}
\end{itemize}

We also use the \textit{Transformers} \cite{wolf-etal-2020-transformers} and \textit{PyTorch Lightning} \cite{pytorch-lightning,pytorch} libraries for Python.
We use the following hyperparameters for finetuning the models:

\begin{table}[h]
\centering
\begin{tabular}{lll}\toprule
Parameter & Grid search & Used \\\midrule
Batch size & 16, 32 & 32 \\
Learning rate & 3e-5, 2e-5 & 2e-5 \\
Epochs & 1, 2, 3 & 2 \\
Classifier dropout & (0.1) & 0.1 \\\bottomrule
\end{tabular}
\caption{\textbf{Hyperparameters used during the grid search and for the final experiments.}}
\label{tab:hyperparam}
\end{table}

\vfill\break

\section{Additional Correlations}
\label{sec:appendix-correlations}

\begin{table}[h]
\begin{adjustbox}{max width=\columnwidth, center}
% workaround for arxiv
% \input{tables/corr-seen-subwords.tex}
\includegraphics[width=\textwidth]{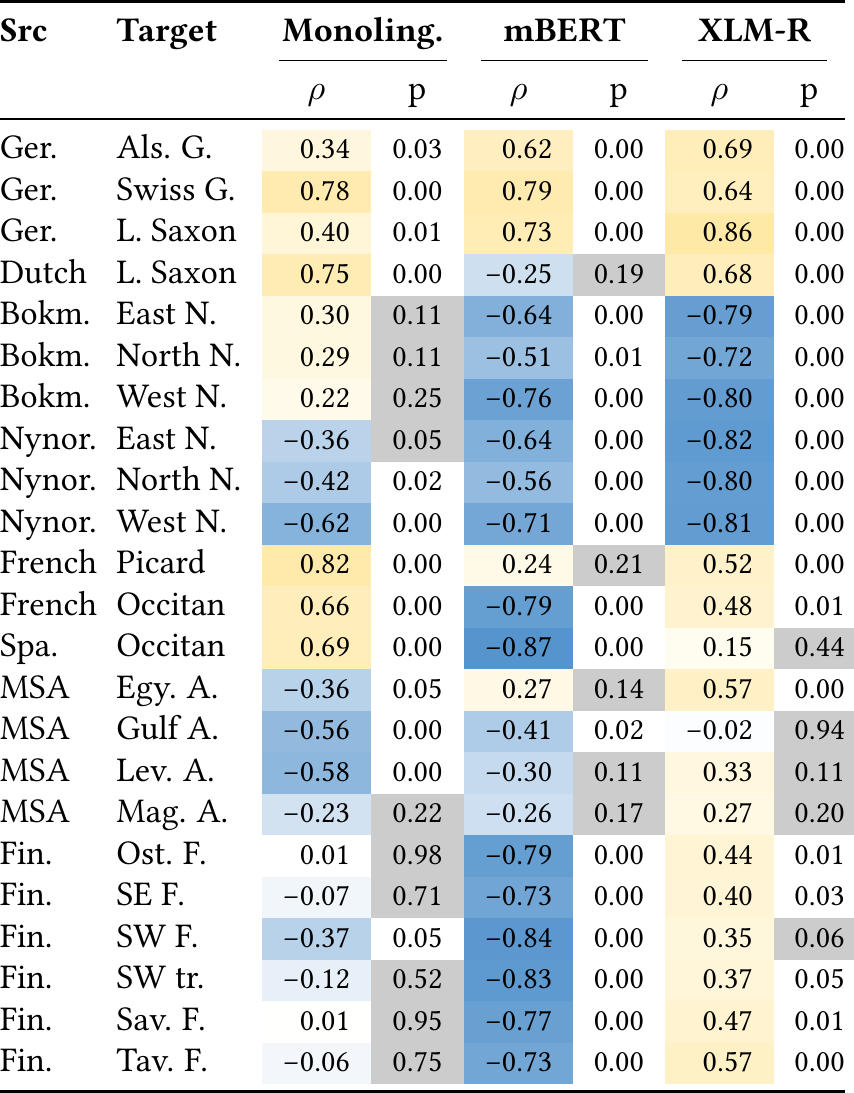}
\end{adjustbox}
\caption{\textbf{Correlation between seen subword ratio and accuracy.}
Spearman's $\rho$ with \textit{p}-values for all noise levels and random initializations per language pair and PLM.
Negative correlations are highlighted in blue, positive ones in yellow.
\textit{P}-values of 0.05 and above have a grey background.
}
\label{tab:corr-seen-subwords}
\end{table}

\begin{table}
\begin{adjustbox}{max width=\columnwidth, center}
% workaround for arxiv
% \input{tables/corr-seen-words.tex}
\includegraphics[width=\textwidth]{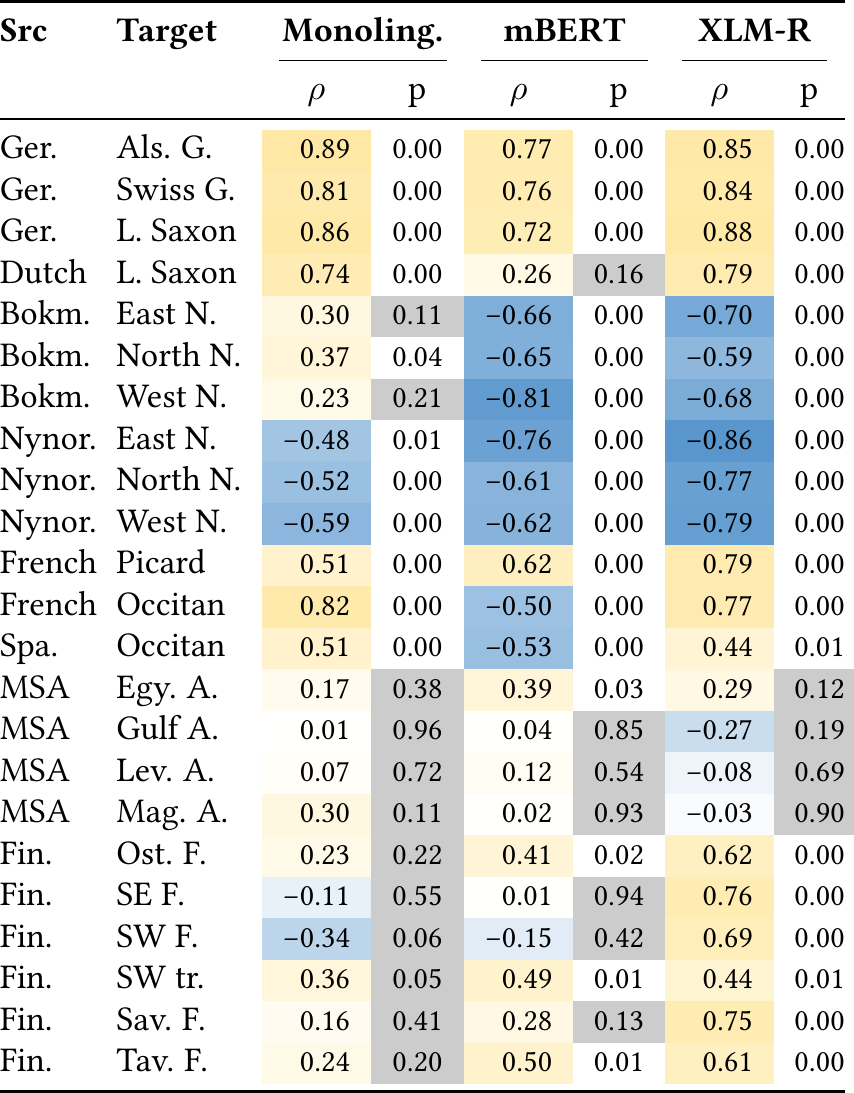}
\end{adjustbox}
\caption{\textbf{Correlation between seen word ratio and accuracy.}
Spearman's $\rho$ with \textit{p}-values for all noise levels and random initializations per language pair and PLM.
Negative correlations are highlighted in blue, positive ones in yellow.
\textit{P}-values of 0.05 and above have a grey background.
}
\label{tab:corr-seen-words}
\end{table}

\begin{table}
\vspace*{24pt} % make the tables appear aligned across columns
\begin{adjustbox}{max width=\columnwidth, center}
% workaround for arxiv
% \input{tables/corr-ttr-ratio.tex}
\includegraphics[width=\textwidth]{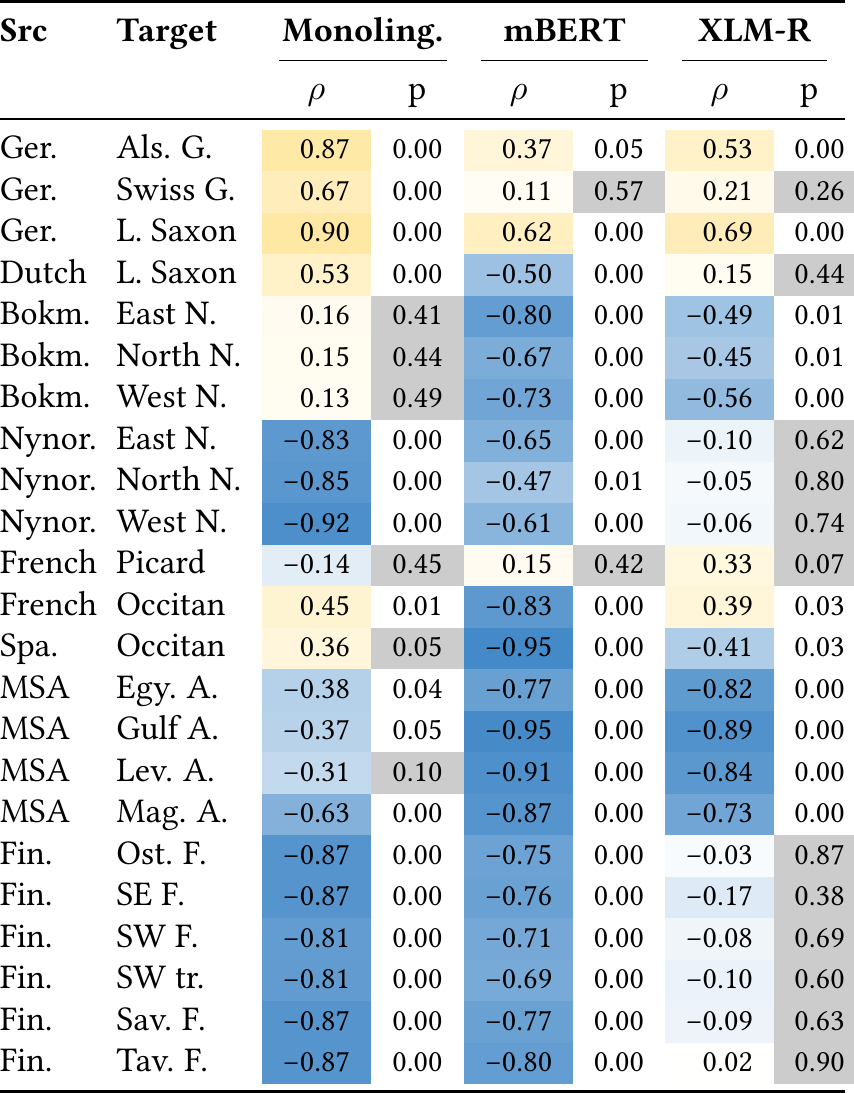}
\end{adjustbox}
\caption{\textbf{Correlation between TTR ratio and accuracy.}
Spearman's $\rho$ with \textit{p}-values for all noise levels and random initializations per language pair and PLM.
Negative correlations are highlighted in blue, positive ones in yellow.
\textit{P}-values of 0.05 and above have a grey background.
The TTR ratio stayed below~$1$ for all cross-dialectal Finnish set-ups (regardless of PLM choice) and above~$1$ for all others.
}
\label{tab:corr-ttr-ratio}
\end{table}

\end{document}